\documentclass[a4paper,fleqn]{cas-dc}

\usepackage[numbers,sort&compress]{natbib}
\usepackage{graphicx}
\usepackage{subcaption}

\def\tsc#1{\csdef{#1}{\textsc{\lowercase{#1}}\xspace}}
\tsc{WGM}
\tsc{QE}
\begin{document}
\let\WriteBookmarks\relax
\def\floatpagepagefraction{1}
\def\textpagefraction{.001}

\shorttitle{}    

\shortauthors{}  

\title [mode = title]{Foundation-Style Pretraining for Scalable and Transferable Distributed PV Forecasting}  

% % Title footnote mark
% % eg: \tnotemark[1]

% % Title footnote 1.
% % eg: \tnotetext[1]{Title footnote text}

% Authors
\author[1,2,3]{Dayin Chen}
\ead{23038748r@connect.polyu.hk}

\author[4]{Xiaodan Shi}
\cormark[1]
\ead{xiaodan.shi@dsv.su.se}

\author[1,2,3]{Xin Xu}
\ead{xin666.xu@connect.polyu.hk}

\author[5]{Zhongyue Su}
\ead{zsu@idt.eitech.edu.cn}

\author[1,2,3]{Shibo Zhu}
\ead{shibo.zhu@connect.polyu.hk}

\author[1,2,3]{Ruizhe Deng}

\author[5]{Dongxiao Zhang}
\cormark[2]
\ead{dzhang@eitech.edu.cn}

\author[5]{Yuntian Chen}
\ead{ychen@eitech.edu.cn}

\author[1,2]{Jinyue Yan}
\cormark[3]
\ead{j-jerry.yan@polyu.edu.hk}

% Affiliations
\affiliation[1]{organization={Department of Building Environment and Energy Engineering, The Hong Kong Polytechnic University},
            city={Hong Kong SAR},
            country={China}}

\affiliation[2]{organization={International Centre of Urban Energy Nexus, The Hong Kong Polytechnic University},
            addressline={Hung Hom, Kowloon},
            state={Hong Kong SAR},
            country={China}}

\affiliation[3]{organization={Ningbo Institute of Digital Twin, Eastern Institute of Technology},
            city={Ningbo},
            postcode={315200},
            country={China}}

\affiliation[4]{organization={Department of Computer and Systems Sciences, Stockholm University},
            city={Stockholm},
            country={Sweden}}

\affiliation[5]{organization={Zhejiang Key Laboratory of Industrial Intelligence and Digital Twin, Ningbo Institute of Digital Twin, Eastern Institute of Technology},
            city={Ningbo},
            postcode={315200},
            country={China}}

% Corresponding authors
\cortext[1]{Corresponding author}
\cortext[2]{Corresponding author}
\cortext[3]{Corresponding author}

\begin{abstract}
The rapid growth of distributed photovoltaic (PV) generation is increasing the demand for accurate forecasting across large numbers of geographically dispersed stations. Existing approaches are commonly developed for individual sites or specific forecasting horizons, requiring repeated training and maintenance as new stations are integrated and limiting direct reuse at previously unseen sites. To address this scalability gap, this paper presents a domain-specific foundation-style pretraining framework that learns a shared forecasting model from heterogeneous multi-site and multi-horizon PV data. NWP-assisted forecasting is formulated as a sequence-completion problem that unifies historical PV power, meteorological observations, future numerical weather prediction (NWP) forecasts, and time encodings. Three pretraining objectives are designed and systematically evaluated, including forecast-aligned reconstruction, generic reconstruction, and mixed reconstruction. The mixed objective combines the operational structure of PV forecasting with random sequence reconstruction and provides the most balanced performance across forecasting accuracy, degraded NWP conditions, and downstream adaptation. Experiments on 479 distributed PV stations use 340 stations for model development and 139 spatially held-out stations for zero-shot evaluation. Across horizons from 6 hours to 7 days, the multi-site pretrained MR model outperforms representative station-specific and shared baselines, as well as generalist time-series models, including a recent large-scale pretrained foundation model. The pretrained model can also be directly deployed at new PV stations and subsequently adapted to region-specific operational objectives when local data become available, providing a scalable forecasting paradigm for large distributed PV fleets.
\end{abstract}

\begin{highlights}

\item A shared pretrained model enables zero-shot forecasting at unseen PV sites.

\item NWP-assisted PV forecasting is formulated as a sequence-completion task.

\item Three pretraining objectives are evaluated; mixed reconstruction performs best.

\item Multi-site pretraining outperforms station-specific and generalist baselines.

\item The pretrained model adapts to region-specific operational objectives.

\end{highlights}

%\nocite{*}

% Keywords
\begin{keywords}
Photovoltaic power forecasting \sep Distributed PV systems \sep Foundation-style pretraining \sep Multi-horizon forecasting \sep Zero-shot transfer
\end{keywords}

\maketitle

% Main text

\section{Introduction}
\label{sec:introduction}

\begin{figure*}[!t]
\centerline{\includegraphics[width=2\columnwidth]{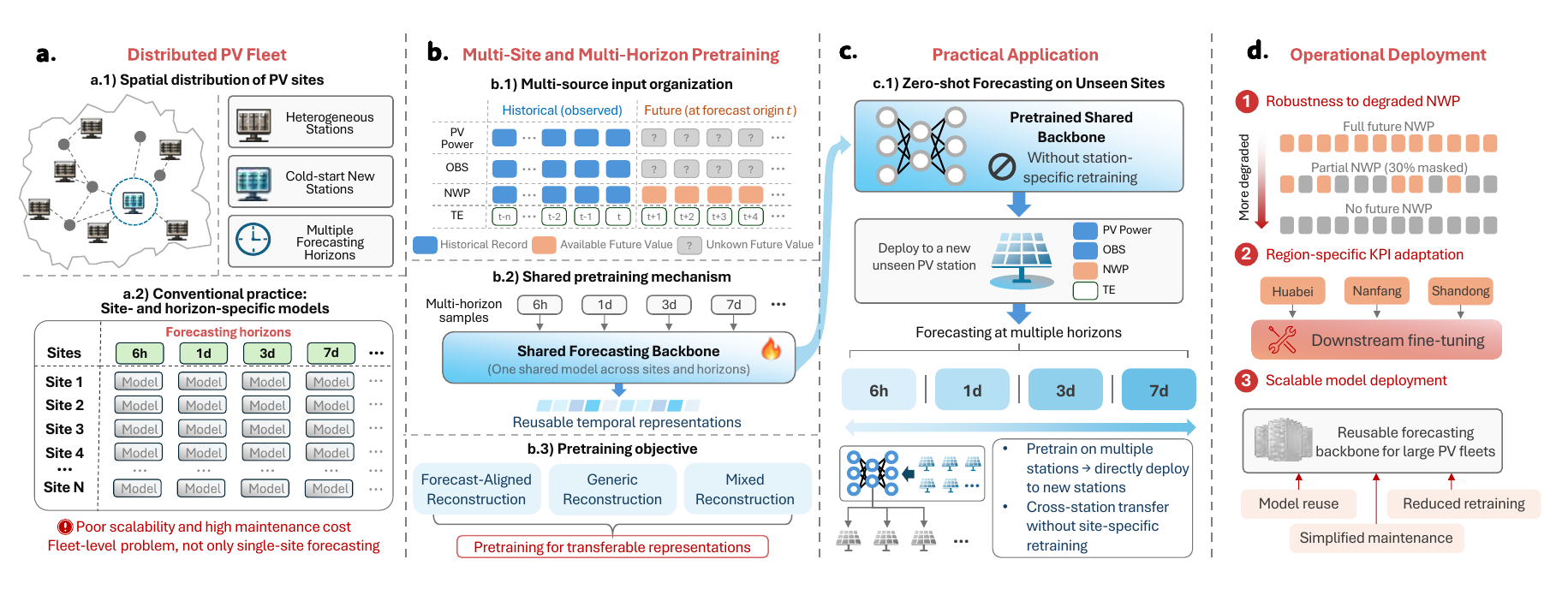}}
\caption{Overview of the proposed foundation-style pretraining framework for scalable distributed PV forecasting. A shared model is pretrained across multiple PV sites and forecasting horizons using a sequence-completion formulation and reconstruction-based objectives, enabling zero-shot transfer to unseen stations, robustness to degraded NWP availability, and downstream adaptation to region-specific operational objectives.}
\label{framework}
\end{figure*}

Solar photovoltaic (PV) deployment is expanding rapidly while becoming increasingly geographically distributed. Global cumulative PV capacity approached 3~TW by the end of 2025, with approximately 698~GW newly installed during the year \cite{R1_old}. In China, distributed PV alone added 153~GW in 2025 and reached a cumulative installed capacity of 530~GW \cite{R2_old}. Unlike utility-scale PV plants, which are typically developed and monitored as spatially concentrated generation units, distributed PV consists of numerous user-side systems deployed across residential, commercial, and industrial settings \cite{DPV_policy,distri_1,distri_2}. These geographically dispersed systems are heterogeneous in installed capacity, operating conditions, and local meteorological environments, resulting in substantial diversity in their generation characteristics \cite{pv_review}. As distributed PV continues to penetrate modern power systems, accurate
forecasting of such heterogeneous resources becomes increasingly important
for power-system dispatch, day-ahead trading, and balancing \cite{usage,lack_data}.

The rapid growth in the number of distributed PV sites also creates a scalability challenge for conventional forecasting pipelines. In many practical applications, forecasting models are still developed and maintained separately for individual stations and prediction horizons \cite{pv_review}. Such station-specific and horizon-specific modeling requires repeated development and maintenance. This effort increases as new PV systems are connected and forecasting requirements change. Transfer learning and multi-site forecasting have therefore been investigated to enable knowledge sharing across PV systems and reduce dependence on site-specific model development \cite{luo2, multisite1,multisite2}. Despite these advances, several limitations remain. Transfer-learning approaches often require adaptation to individual target
stations~\cite{luo2,tang_transfer,sarmas_transfer}, while newly commissioned PV systems may
provide only limited local historical data~\cite{lack_data}. Multi-site forecasting methods, in contrast, are commonly developed for a predefined group of stations \cite{multisite1, multisite2, multisite5}, limiting their direct applicability to previously unseen sites. Moreover, both paradigms are often designed for specific forecasting horizons, meaning that separate models or additional customization may still be required when operational forecasting requirements change. Taken together, these limitations constrain model reuse across heterogeneous and unseen stations and forecasting horizons, motivating a shared pretrained forecasting framework.

Recent advances in pretrained time-series models and foundation-model paradigms have demonstrated the potential of learning reusable temporal representations within a shared backbone \cite{Chronos,timemoe,timesfm3}. Instead of repeatedly training independent models for individual forecasting tasks, such a paradigm enables common temporal patterns to be learned from heterogeneous time series and subsequently reused across datasets, forecasting horizons, and downstream applications. This idea is particularly attractive for large-scale distributed PV systems, where a shared pretrained model could potentially serve many geographically dispersed stations and multiple operational forecasting horizons. In this work, \emph{foundation-style pretraining} refers to a domain-specific pretrain--generalize--adapt paradigm, where a shared forecasting backbone learns reusable representations across multiple PV sites and forecasting horizons.

However, PV forecasting has a distinct operational information structure. At the forecast origin, historical PV generation and meteorological records are available, while future PV generation remains unknown and NWP forecasts over the prediction horizon are already accessible \cite{chen2025selecting}. In this study, these variables are organized according to their forecast-time availability, and NWP-assisted PV forecasting is formulated as a sequence-completion problem. Historical observations and known-future NWP retain their original temporal positions within a unified multivariate sequence, while future PV generation is treated as the missing target. This formulation preserves the temporal alignment among variables and provides a common basis for shared pretraining across multiple PV stations and forecasting horizons.

Building on this formulation, we develop a foundation-style pretraining framework for scalable and transferable distributed PV forecasting. A shared forecasting backbone is trained on variable-horizon samples, allowing a single set of model parameters to support multiple operational forecasting horizons. During pretraining, a mixed reconstruction objective combines forecasting-oriented supervision with auxiliary masked reconstruction to improve cross-site and multi-horizon transferability. The pretrained model can then be directly deployed at unseen PV stations without station-specific retraining, while the pretrained model can be further adapted when specific operational objectives are considered. The overall framework is illustrated in Fig.~\ref{framework}.

The framework is evaluated on a real-world dataset containing 479 distributed PV stations in Ningbo, China, integrating measured PV generation, weather reanalysis data, and rolling NWP forecasts. A strict spatiotemporal evaluation protocol is adopted, in which the shared model is developed using 340 stations and evaluated on 139 previously unseen stations during a temporally separated test period. Experiments cover four forecasting horizons, including 6~hours, 1~day, 3~days, and 7~days. Beyond standard forecasting accuracy, the evaluation examines zero-shot cross-station transfer, robustness under degraded NWP availability, and adaptation to region-specific operational criteria. The results demonstrate competitive forecasting accuracy while enabling a single shared model to be reused across unseen stations and multiple forecasting horizons.

The main contributions of this work are summarized as follows:

\begin{itemize}

\item We develop a foundation-style pretraining framework for scalable distributed PV forecasting, where a single shared model is pretrained across multiple PV sites and forecasting horizons and directly applied to previously unseen stations without station-specific retraining.

\item We formulate NWP-assisted PV forecasting as a sequence-completion task and design three pretraining objectives. Mixed reconstruction achieves the best balance between forecasting accuracy and robustness to degraded NWP availability.

\item We conduct a large-scale evaluation on 479 distributed PV stations under a strict spatiotemporal protocol, including 139 spatially held-out stations. The experiments cover multi-horizon zero-shot forecasting, cross-station generalization, NWP robustness, and downstream adaptation to region-specific operational objectives.

\end{itemize}

\section{Related Work}

\subsection{PV Forecasting with Weather Information}

Weather information has long played a central role in PV power forecasting. Early studies commonly formulated the problem as an NWP-driven regression task, where forecasted irradiance, temperature, humidity, and cloud cover were used as major predictive inputs. Luo et al. developed a physics-constrained LSTM for day-ahead PV forecasting using forecasted meteorological variables \cite{luo1,luo2}. Böök and Lindfors proposed a site-specific post-processing framework that combined weather forecasts, a physical PV model, and rolling historical-output correction to reduce systematic bias \cite{new_ref3}. Mayer et al. converted European Centre for Medium-Range Weather Forecasts (ECMWF) and Applications of Research to Operations at Mesoscale (AROME) forecasts into PV output through a physical model chain for short-term and day-ahead prediction \cite{new_ref4}. Brester et al. replaced historical NWP archives with weather observations during model development and subsequently validated the trained model using operational forecasts, thereby reducing dependence on archived forecast products \cite{new_ref5}. These studies established the importance of future meteorological information for operational PV forecasting.

With the development of machine learning and deep learning, another line of research has focused on extracting temporal patterns from historical PV generation and multivariate observations. Qu et al. proposed an adversarial domain-adaptation framework for cross-regional distributed PV forecasting under limited target-site data availability \cite{ref1}. Wang et al. developed RCLformer by integrating Crossformer, LSTM, and RevIN for long-horizon forecasting under non-stationary conditions \cite{ref6}. Liu and Gai introduced a patch-based MLP for long-sequence PV forecasting using historical multivariate observations \cite{new_ref1}, while Zhu et al. combined LSTM and XGBoost to capture temporal dependencies and nonlinear relationships in long-term plant measurements \cite{new_ref2}. Such history-driven approaches effectively exploit temporal persistence and local operating patterns. However, their predictive information becomes increasingly limited as the forecasting horizon extends.

More recent studies have therefore combined historical observations with future NWP information. Representative approaches include temporal alignment between historical power sequences and future weather forecasts \cite{nwp_shift}, direct feature fusion of NWP variables with historical inputs \cite{nwp_feature}, and dedicated encoder--decoder structures for meteorological covariates \cite{nwp_embed}. Overall, existing studies provide effective mechanisms for exploiting future meteorological information. In contrast to treating NWP primarily as an additional exogenous input, the present work organizes historical observations, known-future NWP, and unknown future PV within their original temporal positions and formulates forecasting as a sequence-completion task. This formulation is subsequently used as the basis for shared pretraining rather than as an isolated NWP-fusion mechanism.

\subsection{Multi-site and Transferable PV Forecasting}

The increasing number of distributed PV systems has motivated forecasting methods that share information across sites rather than training independent models for every station. Two major directions have emerged: transfer learning for adapting knowledge to target sites and multi-site forecasting for jointly modeling groups of spatially distributed PV systems.

Transfer-learning approaches aim to reduce the dependence on target-site training data by reusing representations or model parameters learned from related PV systems. Qu et al. investigated adversarial domain adaptation for cross-regional distributed PV forecasting under limited local data \cite{ref1}. Tang et al. transferred a pretrained forecasting model to newly built PV sites through parameter fine-tuning \cite{tang_transfer}, while Sarmas et al. systematically examined several weight-transfer and fine-tuning strategies for solar power forecasting under data scarcity \cite{sarmas_transfer}. These studies demonstrate the value of knowledge transfer for newly commissioned or data-limited PV systems. Nevertheless, deployment generally still requires target-specific adaptation or access to a certain amount of target-site data.

Multi-site forecasting methods instead exploit information from multiple PV systems during model training. Wen et al. developed RAD-FEDformer, which jointly trains data from multiple PV sites and incorporates geographical correlations through a relevance-amplification module \cite{multisite1}. Simeunovic et al. proposed spatio-temporal graph neural network models, including GCLSTM and GCTrafo, to capture spatial dependencies among hundreds of PV systems \cite{multisite2}. Yang et al. introduced DEST-GNN to model dynamic inter-site relationships through adaptive graph learning \cite{multisite5}. Other studies have explored ensemble-based and multi-objective learning frameworks for improved cross-site generalization \cite{cross_new_1}, while diffusion-based data augmentation has been investigated to increase robustness under heterogeneous site conditions \cite{cross_new_2}. Many of these approaches are constructed around a predefined collection of stations or rely on explicit spatial modeling and additional task-specific modules. Their direct reuse can therefore become less straightforward when new stations are introduced or the forecasting configuration changes.

These limitations motivate a complementary direction based on parameter sharing, where a common forecasting model can be learned from heterogeneous PV systems and reused across stations and operational forecasting horizons.

\subsection{Pretrained and Generalist Time-Series Forecasting}

Recent advances in large-scale time-series modeling have shifted forecasting from task-specific models toward pretrained models with reusable temporal representations. Chronos \cite{Chronos} has become a representative general-purpose foundation-model baseline, while more recent models such as TimeMoE \cite{timemoe} and the TimesFM family \cite{TimesFM} have further advanced large-scale pretrained forecasting. The latest TimesFM-3 \cite{timesfm3} further advances this line of large-scale foundation forecasting. Similar developments have also appeared in weather and climate forecasting \cite{weather_FM}. These models provide an important reference for distributed PV forecasting, but their pretraining is designed to capture broadly transferable patterns across diverse domains rather than characteristics specific to distributed PV fleets. A key challenge therefore remains to learn a shared PV forecasting model from multiple stations and forecasting horizons that can be directly reused at unseen sites. This work addresses this challenge through a domain-specific foundation-style pretraining paradigm tailored to distributed PV forecasting.

\section{Methodology}

\subsection{Shared Pretraining Setting}
\label{sec:shared_pretraining}

We consider a fleet of distributed PV stations with heterogeneous generation characteristics and multiple operational forecasting horizons. Let $\mathcal{S}_{\mathrm{tr}}$ denote the set of stations available during pretraining and $\mathcal{H}$ denote the set of forecasting horizons. For each station $i \in \mathcal{S}_{\mathrm{tr}}$ and horizon $h \in \mathcal{H}$, forecasting samples are constructed from the corresponding PV and meteorological data. Instead of training separate station-specific or horizon-specific models, a single forecasting model $f_{\theta}$ with shared parameters $\theta$ is optimized jointly over samples from different stations and horizons.

Let $\mathcal{D}_{i,h}$ denote the set of training samples associated with station $i$ and forecasting horizon $h$. The shared model is learned through
\begin{equation}
\theta^{*}
=
\arg\min_{\theta}
\sum_{i \in \mathcal{S}_{\mathrm{tr}}}
\sum_{h \in \mathcal{H}}
\mathbb{E}_{(\mathbf{x},\mathbf{y}) \sim \mathcal{D}_{i,h}}
\left[
\mathcal{L}_{\mathrm{pre}}
\left(
f_{\theta}(\mathbf{x}),\mathbf{y}
\right)
\right],
\end{equation}
where $\mathcal{L}_{\mathrm{pre}}$ denotes the reconstruction-based pretraining objective introduced in Sec.~\ref{sec:pretraining_objective}. The same parameter set is shared across all training stations and forecasting horizons. No station identifiers, geographical embeddings, or horizon-specific embeddings are provided to the model. The shared backbone therefore remains independent of any predefined station set or spatial topology and learns station and horizon variations directly from the input sequences. This station-agnostic design enables the same pretrained model to be applied to unseen PV stations without introducing station-specific parameters or modifying the model architecture.

Within this setting, \emph{foundation-style pretraining} refers to the joint learning of a reusable base model from heterogeneous multi-site and multi-horizon data. The pretrained model is then directly applied to a disjoint set of PV stations that are never observed during pretraining. For an unseen station $j \notin \mathcal{S}_{\mathrm{tr}}$, the learned parameters $\theta^{*}$ are used without station-specific retraining, forming the zero-shot deployment setting considered in this study.

Beyond zero-shot deployment, the pretrained model can also serve as an initialization for downstream adaptation when target-station data and specific operational objectives are available. The overall framework therefore follows a pretrain--generalize--adapt workflow, combining shared pretraining, zero-shot generalization to unseen stations, and optional downstream fine-tuning.

A key characteristic of PV forecasting is that future NWP is available at the forecast origin, whereas future PV generation remains unknown. The following subsection formulates this partially observed future context as a sequence-completion problem.

\subsection{Sequence-Completion Formulation}
\label{sec:sequence_completion}

The forecasting framework considers four groups of variables: PV power, meteorological observations (OBS), numerical weather prediction (NWP) forecasts, and deterministic time encodings (TE). In this study, the OBS variables are represented by ERA5 weather reanalysis. The variables included in each group are summarized in Table~\ref{tab:features}.

\begin{table}[!t]
\centering
\caption{Variables included in the multivariate forecasting sequence, grouped into four categories. ERA5 weather reanalysis is used as a proxy for meteorological observations (OBS).}
\label{tab:features}
\renewcommand{\arraystretch}{1.15}

\resizebox{\columnwidth}{!}{%
\begin{tabular}{p{2.3cm} p{5.2cm}}
\hline
\textbf{Category} & \textbf{Variables} \\
\hline

PV power &
PV power output (elec) \\
\hline

Meteorological observations (OBS) &
Relative humidity (rh); 
10-m wind components ($u_{10}$, $v_{10}$);
2-m air temperature ($t_{2m}$);
surface solar radiation downwards (ssrd);
surface thermal radiation downwards (strd);
top-of-atmosphere solar radiation (tsr);
total cloud cover (tcc) \\
\hline

Numerical weather prediction (NWP) &
Forecasted global horizontal irradiance (ghi\_f);
surface pressure (sp\_f);
relative humidity (rh\_f);
2-m air temperature (t2m\_f);
10-m wind direction (wd10\_f);
10-m wind speed (ws10\_f);
total cloud cover (tcc\_f) \\
\hline

Time encoding (TE) &
Hour of day (sine, cosine);
minute of hour (sine, cosine);
day of year (sine, cosine) \\
\hline

\end{tabular}%
}
\end{table}

For a given forecast origin, all variables are organized according to their original temporal positions within a unified multivariate sequence. Let
\begin{equation}
\mathbf{X}
=
\left[
\mathbf{X}^{\mathrm{hist}};
\mathbf{X}^{\mathrm{fut}}
\right]
\in
\mathbb{R}^{S\times C},
\end{equation}
where $S=S_h+S_f$ is the total sequence length, $S_h$ and $S_f$ denote the historical and future segments, respectively, and $C$ is the total number of variables.

All four variable groups are available in the historical segment. Their availability differs in the future segment: future PV generation and meteorological observations are unavailable at the forecast origin, whereas future NWP forecasts and time encodings remain available. The resulting forecast-time information structure can therefore be summarized as

\begin{equation}
\begin{array}{c|cccc}
 & \mathrm{PV} & \mathrm{OBS} & \mathrm{NWP} & \mathrm{TE} \\
\hline
\mathrm{History} & \checkmark & \checkmark & \checkmark & \checkmark \\
\mathrm{Future}  & \times     & \times     & \checkmark & \checkmark
\end{array}.
\end{equation}

Based on this forecast-time availability pattern, an operational availability mask is defined as
\begin{equation}
\mathbf{A}\in\{0,1\}^{S\times C},
\end{equation}
where $A_{t,c}=1$ indicates that variable $c$ at time $t$ is available at the forecast origin, and $A_{t,c}=0$ otherwise. The partially observed sequence is then constructed as
\begin{equation}
\widetilde{\mathbf{X}}
=
\operatorname{Mask}\left(\mathbf{X},\mathbf{A}\right),
\end{equation}
where $\operatorname{Mask}(\cdot)$ constructs the partially observed sequence according to the availability mask $\mathbf{A}$.

Under this representation, NWP-assisted PV forecasting is formulated as a sequence-completion task. Given the partially observed sequence $\widetilde{\mathbf{X}}$, the shared forecasting model reconstructs the multivariate sequence as
\begin{equation}
\widehat{\mathbf{X}}
=
f_{\theta}\left(\widetilde{\mathbf{X}}\right),
\end{equation}
where $f_{\theta}$ denotes the forecasting model with parameters $\theta$, and $\widehat{\mathbf{X}}$ is the reconstructed sequence. The final PV forecast is obtained from the future PV component,
\begin{equation}
\widehat{\mathbf{Y}}
=
\widehat{\mathbf{X}}_{\mathrm{PV}}^{\mathrm{fut}}.
\end{equation}

By organizing historical information and future NWP forecasts within a unified temporal sequence, the sequence-completion formulation provides a common representation across PV stations and forecasting horizons. The construction of multi-horizon forecasting samples and forecast-time NWP alignment is described next.

\subsection{Multi-Horizon Sample Construction and Forecast-Time NWP Alignment}
\label{sec:sample_construction}

To support multiple forecasting horizons within a single shared model, training samples are constructed with variable prediction lengths. For each sample, the forecasting horizon $F$ is selected from
\begin{equation}
F \in \{24,40,120,280\},
\end{equation}
corresponding to forecasting tasks of 6 hours, 1 day, 3 days, and 7 days, respectively. Since only the daytime operating period from 07{:}00 to 17{:}00 is retained, one operational day contains 40 samples at the 15-minute resolution.

A common sequence structure is adopted to enable multi-horizon forecasting within a fixed-length model architecture. The historical and future segments are both fixed to
\begin{equation}
S_h=S_f=280,
\end{equation}
giving a total sequence length of $S=560$. For a forecasting horizon $F<S_f$, only the first $F$ positions of the future segment constitute the valid forecasting interval, while the remaining positions are padded and excluded from supervision. This design allows samples with different forecasting horizons to share the same input dimensions and be processed by a single forecasting backbone.

\begin{figure}[!t]
\centering
\includegraphics[width=\columnwidth]{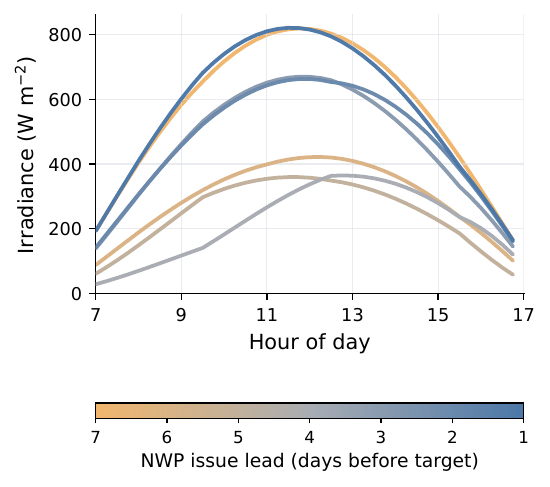}
\caption{Illustration of rolling NWP alignment for the same target day. Solar irradiance forecasts for day $d$ are shown using NWP products issued from days $d-7$ to $d-1$.}
\label{fig}
\end{figure}

NWP forecasts are aligned with the forecast origin of each sample. Let $t$ denote the forecast origin. For a sample with forecasting horizon $F$, the NWP inputs are selected from the forecast product available at time $t$ and covering the corresponding future interval
\begin{equation}
[t+1,t+F].
\end{equation}
As NWP products are updated over time, forecasts for the same target period may differ according to their issue time. The rolling NWP construction therefore associates each forecasting sample with the meteorological forecast that would have been available when the PV forecast was issued. An example of this rolling alignment is illustrated in Fig.~\ref{fig}.

\subsection{Shared Forecasting Backbone}
\label{sec:backbone}

PatchTSMixer \cite{patchtsmixer} is used as the base backbone of the proposed forecasting framework to process the multivariate sequences constructed above.

Given the partially observed sequence
$\widetilde{\mathbf{X}}\in\mathbb{R}^{S\times C}$,
the temporal dimension is divided into non-overlapping patches of length $p$. The number of patches is
\begin{equation}
N=\frac{S}{p},
\end{equation}
and the resulting patch representation can be written as
\begin{equation}
\mathbf{X}_{p}\in\mathbb{R}^{N\times p\times C}.
\end{equation}
Each patch is projected into a latent representation before being processed by the mixer backbone.

The backbone consists of $K$ stacked mixer blocks. Each block performs feature mixing along the temporal-patch, within-patch, and variable dimensions. These operations capture temporal dependencies across patches, local variations within individual patches, and interactions among PV, OBS, NWP, and time-encoding variables. Residual connections and normalization are applied throughout the mixer blocks to facilitate representation learning over the long input sequence.

The output of the final mixer block is mapped by a reconstruction head to the original sequence space,
\begin{equation}
\widehat{\mathbf{X}}
=
g_{\phi}\!\left(
h_{\theta}(\widetilde{\mathbf{X}})
\right),
\end{equation}
where $h_{\theta}(\cdot)$ denotes the shared PatchTSMixer backbone and $g_{\phi}(\cdot)$ denotes the reconstruction head. The reconstructed future PV component is used as the forecasting output according to the sequence-completion formulation in Sec.~\ref{sec:sequence_completion}.

\begin{figure*}[!t]
\centerline{\includegraphics[width=2\columnwidth]{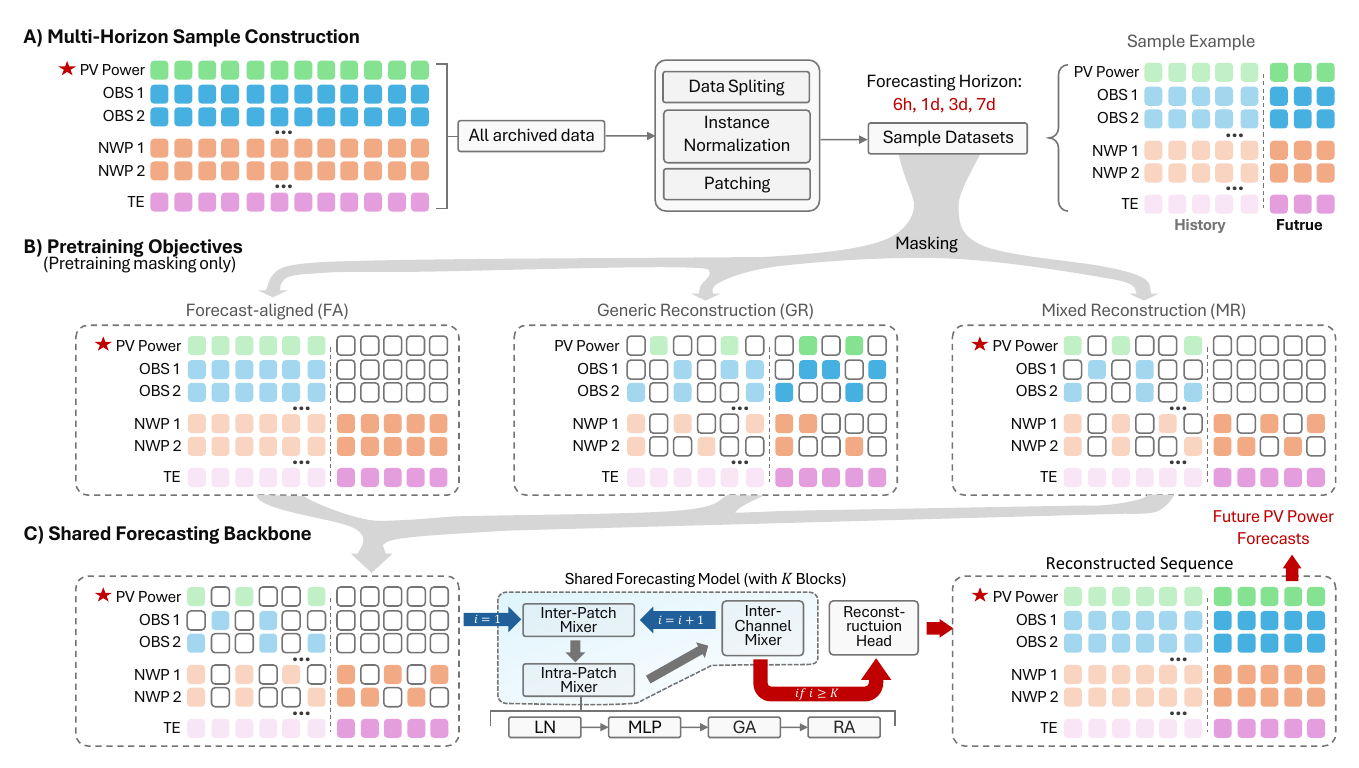}}
\caption{Overview of the multi-horizon sample construction, reconstruction-based pretraining objectives, and shared forecasting backbone. FA applies forecasting-aligned reconstruction, GR performs generic reconstruction with random masking, and MR combines the two masking strategies. The resulting masked sequences are processed by a shared PatchTSMixer-based backbone and reconstruction head, from which the future PV segment is extracted as the forecasting output. White blocks denote masked entries. Pretraining masks are used only during pretraining.}

\label{variants}
\end{figure*}

\begin{figure}[!t]
\centerline{\includegraphics[width=\columnwidth]{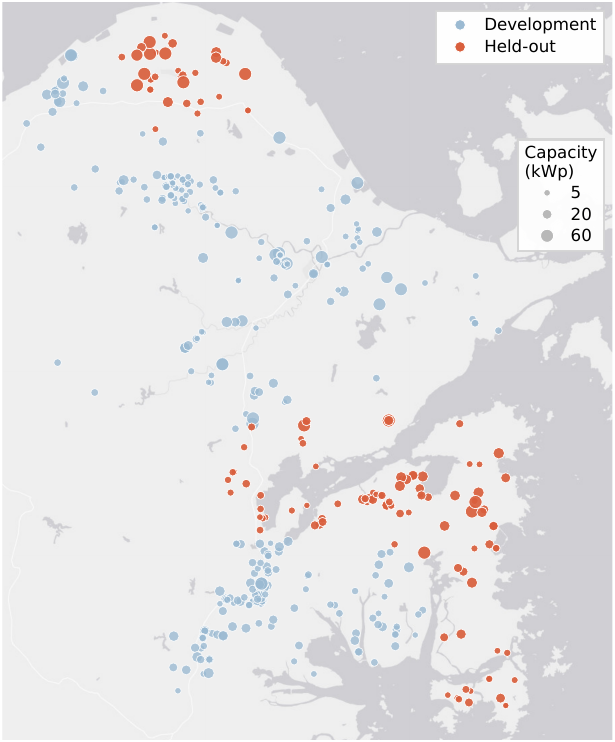}}
\caption{Spatial distribution and dataset split of the 479 distributed PV stations in Ningbo. Colors distinguish the 340 development stations from the 139 spatially held-out stations used for zero-shot evaluation, while marker size represents the installed PV capacity of each station.}

\label{fig:map}
\end{figure}

\subsection{Reconstruction-Based Pretraining Objectives}
\label{sec:pretraining_objective}

Based on the sequence-completion formulation, different pretraining objectives can be constructed by varying the masking pattern and the positions used for reconstruction supervision. These masking schemes are applied during pretraining to expose the model to different partial-observation patterns and encourage transferable representation learning. Three objectives are considered in this work: forecast-aligned reconstruction (FA), generic reconstruction (GR), and mixed reconstruction (MR), as illustrated in Fig.~\ref{variants}.

Let $\mathcal{M}_q$ denote the set of masked positions under objective $q\in\{\mathrm{FA},\mathrm{GR},\mathrm{MR}\}$, and let $\Omega_q$ denote the corresponding supervision set. Given the masked input, the reconstruction loss is defined as
\begin{equation}
\mathcal{L}_q
=
\frac{1}{|\Omega_q|}
\sum_{(t,c)\in\Omega_q}
\left(
\widehat{X}_{t,c}-X_{t,c}
\right)^2.
\end{equation}

\paragraph{Forecast-Aligned Reconstruction (FA)}
FA follows the forecast-time information structure defined in Sec.~\ref{sec:sequence_completion}. Future PV and OBS values are masked, while historical variables, future NWP forecasts, and future time encodings remain available. The reconstruction loss is computed only over the future PV segment,
\begin{equation}
\Omega_{\mathrm{FA}}
=
\Omega_{\mathrm{PV}}^{\mathrm{fut}},
\end{equation}
thereby aligning the supervision directly with the PV forecasting task.

\paragraph{Generic Reconstruction (GR)}
GR adopts a generic masked-reconstruction objective by randomly masking temporal patches of the non-temporal variables across the sequence, while time-encoding variables remain available. The model reconstructs the randomly masked entries, and the corresponding masked positions are used for supervision. Since the forecast-time availability pattern is not explicitly imposed during GR pretraining, some future PV and OBS positions may remain visible when they are not selected for random masking. By varying the observed and masked regions across training samples, GR encourages the model to infer missing information from diverse temporal and cross-variable contexts rather than learning only the fixed information pattern of the forecasting task.

\paragraph{Mixed Reconstruction (MR)}
MR integrates the forecasting-oriented objective of FA with the generic reconstruction objective of GR. Starting from the FA masking pattern, an additional random mask is applied to all non-temporal variables across the sequence with probability $a$. The resulting masking pattern is therefore given by
\begin{equation}
\mathcal{M}_{\mathrm{MR}}
=
\mathcal{M}_{\mathrm{FA}}
\cup
\mathcal{M}_{\mathrm{rand}}.
\end{equation}
The reconstruction loss is computed over all valid masked positions, including the future PV and OBS targets defined by FA and the additional positions introduced by random masking. Time-encoding variables and padded positions are excluded. This allows MR to learn from both forecasting-oriented supervision and broader temporal and cross-variable reconstruction. The random masking probability is set to $a=0.3$.

The above masking schemes are used only during pretraining. During zero-shot forecasting and downstream deployment, no additional random masking is applied, and the input sequence follows the operational availability pattern defined in Sec.~\ref{sec:sequence_completion}: future PV and OBS values are unavailable, whereas future NWP forecasts and time encodings remain available.

\section{Experiments}

\subsection{Dataset and Evaluation Protocol}
\label{sec:dataset_protocol}

The experiments are conducted on a real-world distributed PV dataset collected from 479 stations in Ningbo, China, covering the period from January 2022 to June 2023. PV power measurements are accompanied by meteorological information from two sources. ERA5 weather reanalysis with a spatial resolution of $0.25^{\circ}$ is used to represent meteorological observations, while NWP forecasts are obtained from 12 predefined reference locations across Ningbo. Each PV station is associated with the nearest NWP reference location according to geographical proximity. The spatial distribution and dataset split of the PV stations are shown in Fig.~\ref{fig:map}, where marker size represents the installed capacity of each station.

All data sources are resampled to a unified 15-minute resolution. Only daytime periods from 07{:}00 to 17{:}00 are retained, resulting in 40 samples per operational day. Missing values are filled using linear interpolation. PV power is normalized by the installed capacity of each station and expressed in per-unit (p.u.) form.

A strict spatial and temporal separation is adopted throughout model development and evaluation. For the spatial split, four NWP reference locations are randomly selected, and all PV stations associated with these locations are assigned to the held-out set. This results in 340 development stations and 139 held-out stations, with no overlap between the two groups. Splitting the dataset at the NWP-reference level prevents neighboring stations that share the same NWP inputs from being distributed across the development and held-out sets. The held-out stations therefore form several spatially concentrated groups, providing a more stringent setting for evaluating cross-station generalization than a station-wise random split.

For the 340 development stations, the data are divided chronologically into training, validation, and test periods. The training period spans January 1, 2022 to January 13, 2023, followed by validation from January 14 to March 8, 2023 and testing from March 9 to June 24, 2023. For the shared pretrained models, both pretraining and model selection are performed exclusively using the development stations.

The 139 held-out stations are evaluated over the same final test period from March 9 to June 24, 2023. These stations are completely excluded from shared pretraining and shared-model selection. For zero-shot evaluation, the pretrained model is directly applied to these stations without station-specific retraining or parameter adaptation. This protocol therefore evaluates whether a model pretrained on the development stations can generalize directly to spatially unseen PV stations during a temporally future test period.

\subsection{Experimental Settings and Baselines}
\label{sec:experimental_settings}

The PatchTSMixer-based backbone uses a patch length of $p=4$, a hidden dimension of $H=64$, $K=6$ stacked mixer blocks, and an MLP expansion factor of $\gamma=3$. The proposed models are trained using the Adam optimizer with a batch size of 128, an initial learning rate of $2\times10^{-4}$, and a dropout rate of 0.1. Training is performed for up to 20 epochs with early stopping based on validation performance. Unless otherwise stated, the random masking probability used in GR and MR is set to $a=0.3$.

Three groups of baseline methods are considered to evaluate the proposed framework from complementary perspectives.

\textbf{Station-specific baselines} represent conventional forecasting pipelines in which separate models are developed for individual PV stations and forecasting horizons. The compared methods include persistence-based predictors (Pers-L and Pers-D), Ridge regression, LightGBM \cite{lgbm}, XGBoost \cite{xgb}, Transformer \cite{transformer}, and Informer \cite{informer}. For the learnable baselines, a separate model is trained using historical data from each target station for each forecasting horizon. These methods therefore provide locally customized reference models for comparison with the zero-shot shared forecasting framework.

\textbf{Shared forecasting baselines} are introduced to evaluate cross-station parameter sharing under the same development and evaluation protocol as the proposed framework. A shared Transformer is trained jointly on the 340 development stations and subsequently applied to the held-out stations without station-specific retraining.

\textbf{Generalist time-series baselines} include Chronos \cite{Chronos}, its multivariate variant Chronos-MV, and TimesFM-3 \cite{timesfm3}. Chronos and Chronos-MV are retrained on the 340 development stations under the forecasting setting considered in this study and are subsequently evaluated on the held-out stations without target-station adaptation.

TimesFM-3, released by Google Research in August 2026 \cite{timesfm3}, is the latest large-scale time-series foundation model in the TimesFM family and delivers state-of-the-art benchmark performance. It contains approximately 330 million parameters and is pretrained on a broad mixture of real-world and synthetic time series comprising more than one trillion time points. TimesFM-3 natively supports multivariate forecasting together with past-only and past-and-future covariates, making it particularly suitable for NWP-assisted PV forecasting. Given its large-scale cross-domain pretraining, the official pretrained checkpoint is directly adopted in this study without retraining or fine-tuning on the development stations. Historical PV power is used as the target series, historical meteorological observations as past-only covariates, and NWP together with time encodings as past-and-future covariates.

To reduce the influence of model capacity where practical, the shared Transformer, Chronos, and the proposed model are configured with comparable parameter counts of approximately one million parameters. Chronos-MV is retained as a moderately larger multivariate reference, whereas TimesFM-3 is evaluated at its original pretrained scale. A common model-independent quality-control procedure is applied before evaluation, and all methods are compared on the same valid test samples.

\subsection{Evaluation Metrics}
\label{sec:evaluation}

Forecasting performance is evaluated using standard statistical errors and region-specific operational metrics. The two types of metrics are computed in different representations. MAE and RMSE are evaluated using capacity-normalized PV power in per-unit (p.u.) form, whereas the regional KPIs are calculated after restoring the predictions and observations to the physical power scale of each station.

\subsubsection{Statistical Metrics}

Let $\{y_t\}_{t=1}^{N}$ and $\{\hat{y}_t\}_{t=1}^{N}$ denote the ground-truth and predicted PV power in p.u., respectively. The mean absolute error (MAE) is defined as
\begin{equation}
\mathrm{MAE}
=
\frac{1}{N}
\sum_{t=1}^{N}
\left|y_t-\hat{y}_t\right|,
\end{equation}
and the root mean squared error (RMSE) is
\begin{equation}
\mathrm{RMSE}
=
\sqrt{
\frac{1}{N}
\sum_{t=1}^{N}
\left(y_t-\hat{y}_t\right)^2
}.
\end{equation}
Both metrics are dimensionless under the p.u. representation, with lower values indicating better forecasting accuracy.

\subsubsection{Region-Specific Operational Metrics}

For the regional KPI evaluation, the predicted and observed PV values are first converted from p.u. values back to the physical power scale of each station. Let $\{P_t\}_{t=1}^{N}$ and $\{\hat{P}_t\}_{t=1}^{N}$ denote the corresponding observed and predicted power, and let $\mathrm{cap}$ denote the installed capacity of the station in the same physical unit.

\paragraph{Nanfang KPI}
For the Nanfang region, low-output periods are excluded using a threshold $\tau\,\mathrm{cap}$. The evaluated time index set is
\begin{equation}
\mathcal{T}
=
\left\{
t \,\middle|\,
P_t \ge \tau\,\mathrm{cap}
\ \text{or}\
\hat{P}_t \ge \tau\,\mathrm{cap}
\right\}.
\end{equation}
The corresponding KPI is
\begin{equation}
\mathrm{KPI}_{\mathrm{NF}}
=
\max\!\left(
0,\,
1-
\sqrt{
\frac{1}{|\mathcal{T}|}
\sum_{t\in\mathcal{T}}
\frac{
\left(P_t-\hat{P}_t\right)^2
}{
\max\!\left(
P_t^2,\,
(\delta\,\mathrm{cap})^2
\right)
}
}
\right),
\end{equation}
where $\tau=0.1$ and $\delta=0.2$.

\paragraph{Shandong KPI}
For the Shandong region, forecasting error is evaluated using capacity-normalized RMSE:
\begin{equation}
\mathrm{KPI}_{\mathrm{SD}}
=
1-
\frac{
\sqrt{
\frac{1}{N}
\sum_{t=1}^{N}
\left(P_t-\hat{P}_t\right)^2
}
}{
\mathrm{cap}
}.
\end{equation}

\paragraph{Huabei KPI}
For the Huabei region, larger point-wise forecasting errors receive higher weights. The normalized error weight is
\begin{equation}
w_t
=
\frac{
\left|P_t-\hat{P}_t\right|
}{
\sum_{k=1}^{N}
\left|P_k-\hat{P}_k\right|
},
\end{equation}
and the corresponding KPI is
\begin{equation}
\mathrm{KPI}_{\mathrm{HB}}
=
1-
\frac{
\sqrt{
\sum_{t=1}^{N}
w_t
\left(P_t-\hat{P}_t\right)^2
}
}{
\mathrm{cap}
}.
\end{equation}

Higher KPI values indicate better performance under the corresponding regional assessment rule.

\subsection{Zero-Shot Multi-Horizon Forecasting}
\label{sec:zero_shot_results}

\begin{table*}[!t]
\renewcommand{\arraystretch}{1.15}
\caption{Comparison with shared and generalist time-series models on the common quality-controlled test set of the spatially held-out PV stations under different forecasting horizons. Lower values indicate better performance. Parameter counts are approximately matched for Shared Transformer, Chronos, and the proposed models, while TimesFM-3 is evaluated at its original pretrained scale. All models are evaluated without target-station adaptation.}
\label{tab:global_results}
\centering
\setlength{\tabcolsep}{4pt}
\begin{tabular}{lccccccccc}
\hline
& 
& \multicolumn{2}{c}{6 h}
& \multicolumn{2}{c}{1 day}
& \multicolumn{2}{c}{3 days}
& \multicolumn{2}{c}{7 days} \\
\cline{3-4}\cline{5-6}\cline{7-8}\cline{9-10}
\textbf{Method}
& \textbf{\#Params}
& MAE & RMSE
& MAE & RMSE
& MAE & RMSE
& MAE & RMSE \\
\hline

\multicolumn{10}{l}{\textit{Shared forecasting baseline}} \\
Shared Transformer & 1.07M & 0.124 & 0.161 & 0.122 & 0.159 & 0.125 & 0.163 & 0.131 & 0.170 \\

\multicolumn{10}{l}{\textit{Generalist time-series baselines}} \\
Chronos            & 1.08M & 0.229 & 0.307 & 0.237 & 0.317 & 0.260 & 0.340 & 0.262 & 0.344 \\
Chronos-MV         & 1.52M & 0.130 & 0.196 & 0.161 & 0.237 & 0.208 & 0.292 & 0.226 & 0.309 \\
TimesFM-3          & 330M  & 0.096 & 0.138 & 0.102 & 0.146 & 0.117 & 0.162 & 0.134 & 0.184 \\

\multicolumn{10}{l}{\textit{Proposed framework}} \\
FA (Ours)          & 1.03M & 0.100 & 0.137 & 0.106 & 0.145 & 0.116 & 0.156 & 0.133 & 0.178 \\
MR (Ours)          & 1.03M & \textbf{0.095} & \textbf{0.129} & \textbf{0.100} & \textbf{0.135} & \textbf{0.111} & \textbf{0.147} & \textbf{0.127} & \textbf{0.168} \\
GR (Ours)          & 1.03M & 0.107 & 0.140 & 0.116 & 0.150 & 0.130 & 0.165 & 0.143 & 0.180 \\

\hline
\end{tabular}
\end{table*}

\begin{table*}[!t]
\renewcommand{\arraystretch}{1.15}
\caption{Comparison with conventional station-specific baselines on 139 unseen PV stations under different forecasting horizons. Lower values indicate better performance. The proposed framework is evaluated in a zero-shot transfer setting without station-specific retraining.}
\label{tab:stationwise_results}
\centering
\setlength{\tabcolsep}{4pt}
\begin{tabular}{lcccccccc}
\hline
& \multicolumn{2}{c}{6 h}
& \multicolumn{2}{c}{1 day}
& \multicolumn{2}{c}{3 days}
& \multicolumn{2}{c}{7 days} \\
\cline{2-3}\cline{4-5}\cline{6-7}\cline{8-9}
\textbf{Method}
& MAE & RMSE
& MAE & RMSE
& MAE & RMSE
& MAE & RMSE \\
\hline

\multicolumn{9}{l}{\textit{Station-specific baselines}} \\
Pers-L      & 0.188 & 0.257 & 0.195 & 0.264 & 0.231 & 0.302 & 0.242 & 0.311 \\
Pers-D      & 0.239 & 0.309 & 0.243 & 0.311 & 0.255 & 0.324 & 0.251 & 0.320 \\
Ridge       & 0.105 & 0.141 & 0.107 & 0.143 & 0.116 & 0.154 & 0.131 & 0.172 \\
LGBM        & 0.109 & 0.147 & 0.112 & 0.151 & 0.122 & 0.163 & 0.138 & 0.182 \\
XGB         & 0.108 & 0.146 & 0.111 & 0.149 & 0.121 & 0.161 & 0.137 & 0.179 \\
Transformer & 0.102 & 0.136 & 0.104 & 0.139 & 0.116 & 0.153 & 0.131 & 0.172 \\
Informer    & 0.103 & 0.137 & 0.104 & 0.139 & 0.114 & 0.151 & 0.129 & 0.169 \\

\multicolumn{9}{l}{\textit{Proposed framework}} \\
FA (Ours)   & 0.100 & 0.137 & 0.106 & 0.145 & 0.116 & 0.156 & 0.133 & 0.178 \\
MR (Ours)   & \textbf{0.095} & \textbf{0.129} & \textbf{0.100} & \textbf{0.135} & \textbf{0.111} & \textbf{0.147} & \textbf{0.127} & \textbf{0.168} \\
GR (Ours)   & 0.107 & 0.140 & 0.116 & 0.150 & 0.130 & 0.165 & 0.143 & 0.180 \\

\hline
\end{tabular}
\end{table*}

\begin{figure*}[t]
    \centering
    \begin{subfigure}{0.48\textwidth}
        \centering
        \includegraphics[width=\textwidth]{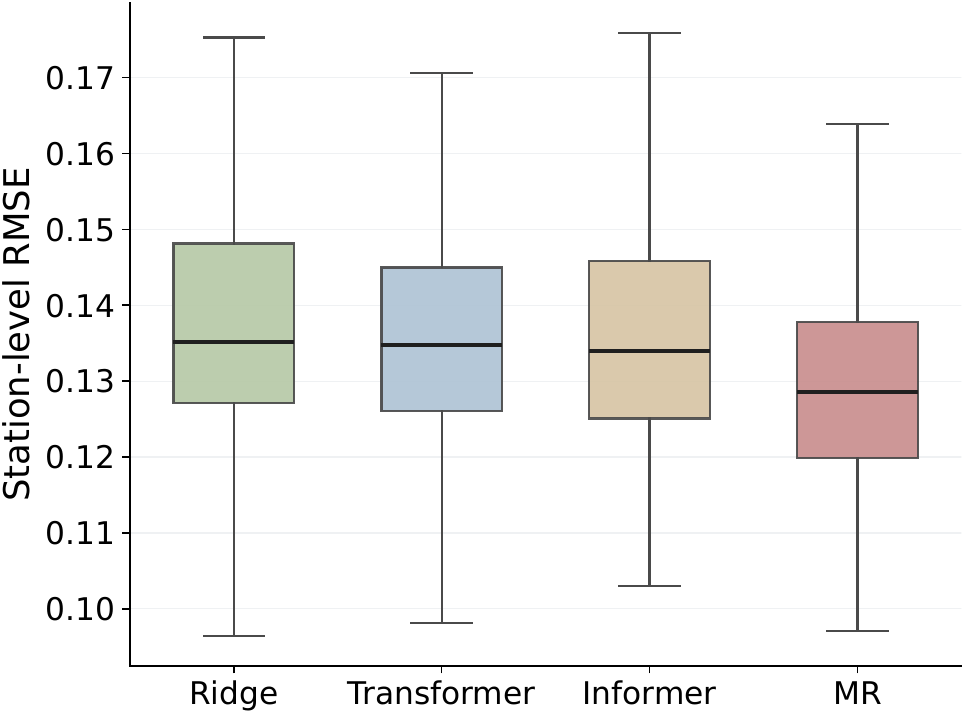}
        \caption{6 h}
        \label{fig:box_h24}
    \end{subfigure}
    \hfill
    \begin{subfigure}{0.48\textwidth}
        \centering
        \includegraphics[width=\textwidth]{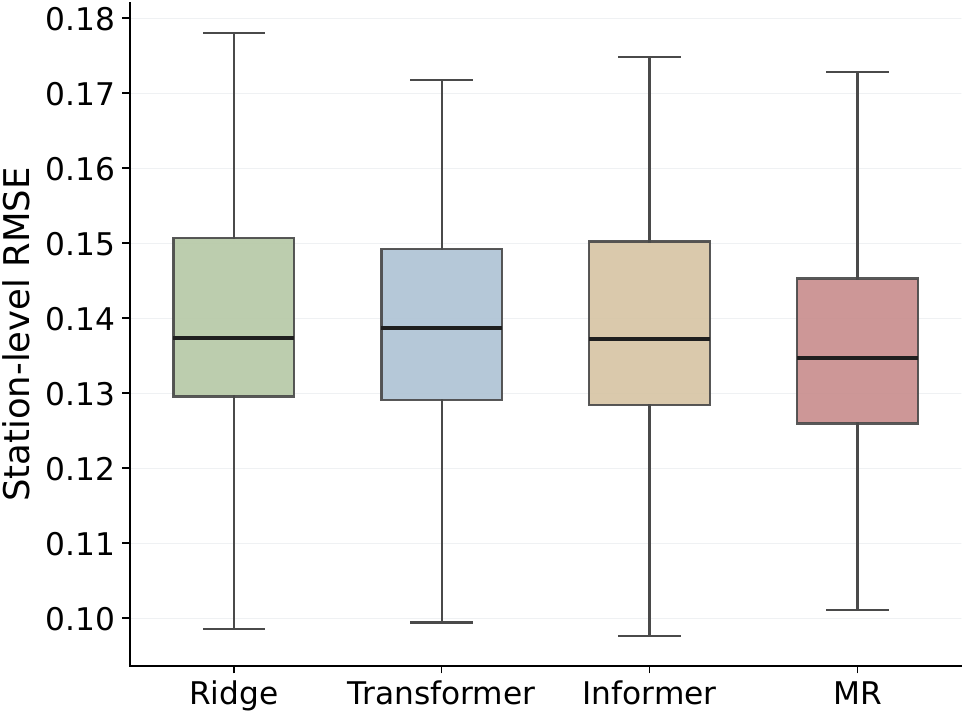}
        \caption{1 day}
        \label{fig:box_h40}
    \end{subfigure}

    \vspace{0.5em}

    \begin{subfigure}{0.48\textwidth}
        \centering
        \includegraphics[width=\textwidth]{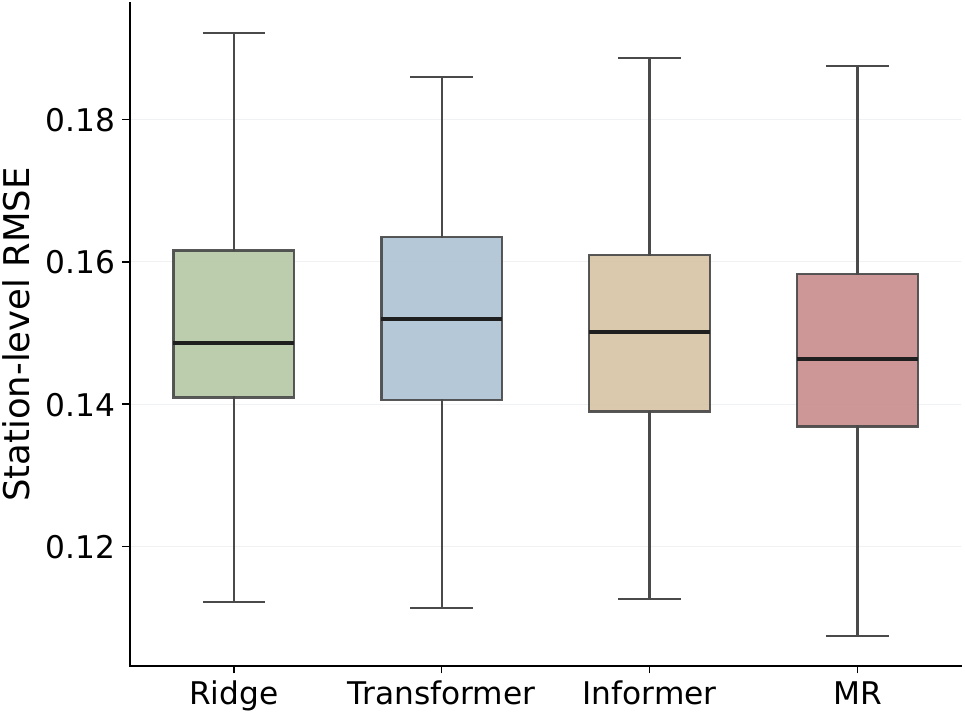}
        \caption{3 days}
        \label{fig:box_h120}
    \end{subfigure}
    \hfill
    \begin{subfigure}{0.48\textwidth}
        \centering
        \includegraphics[width=\textwidth]{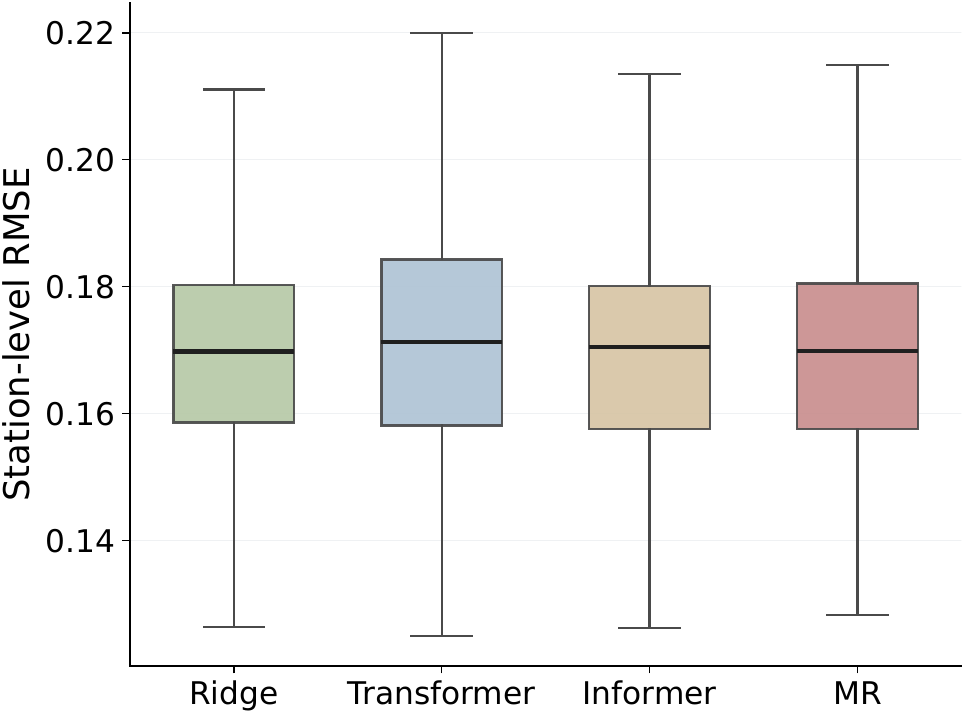}
        \caption{7 days}
        \label{fig:box_h280}
    \end{subfigure}
    \caption{Station-level RMSE distributions of Ridge, Transformer, Informer, and MR across unseen PV stations under four forecasting horizons. Lower values indicate better predictive accuracy.}
    \label{fig:rmse_boxplots_all}
\end{figure*}

% \begin{figure}[!t]
%     \centering
%     \includegraphics[width=0.9\columnwidth]{figs/geo_improvement_MR_vs_Transformer_H40.pdf}
%     \caption{Station-level RMSE improvement of MR over the station-specific Transformer baseline on unseen PV stations under the 1-day forecasting horizon. Positive values indicate lower RMSE achieved by MR, while negative values indicate that Transformer performs better.}
%     \label{fig:geo_improvement_mr_transformer}
% \end{figure}

\begin{figure*}[!t]
    \centering

    \subfloat[$H = 6$ hours]{
        \includegraphics[width=0.48\textwidth]{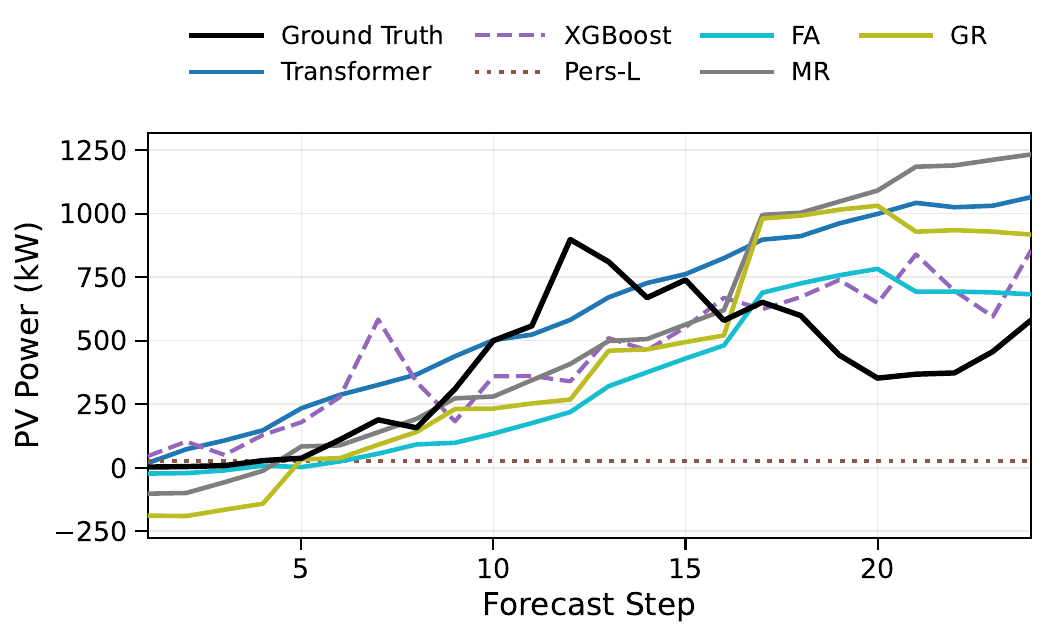}
        \label{fig:qual_case_h24}
    }
    \hfill
    \subfloat[$H = 1$ day]{
        \includegraphics[width=0.48\textwidth]{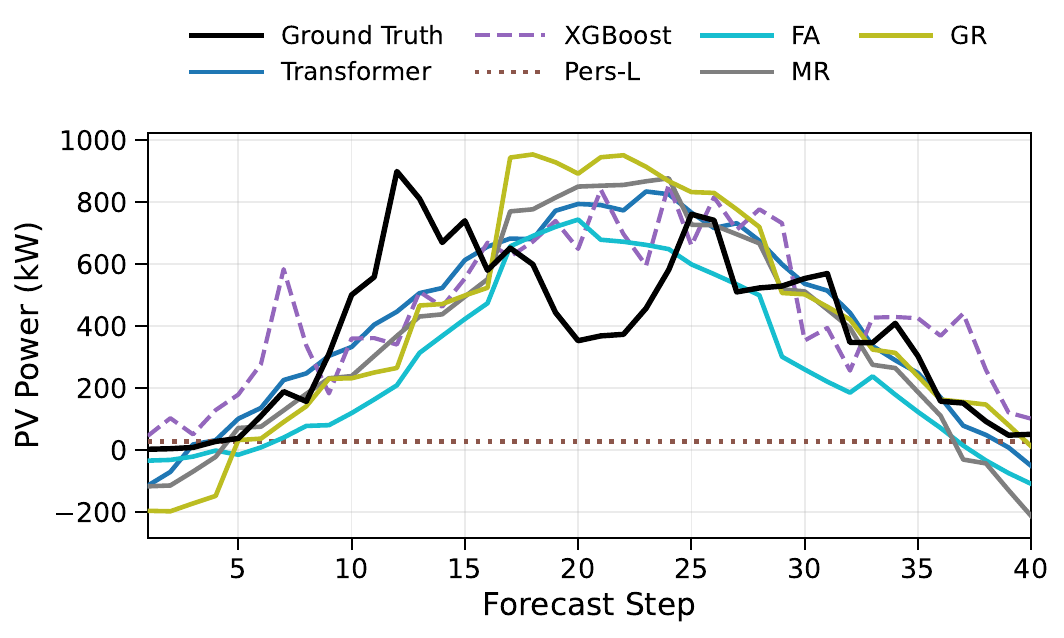}
        \label{fig:qual_case_h40}
    }

    \vspace{0.5em}

    \subfloat[$H = 3$ days]{
        \includegraphics[width=0.48\textwidth]{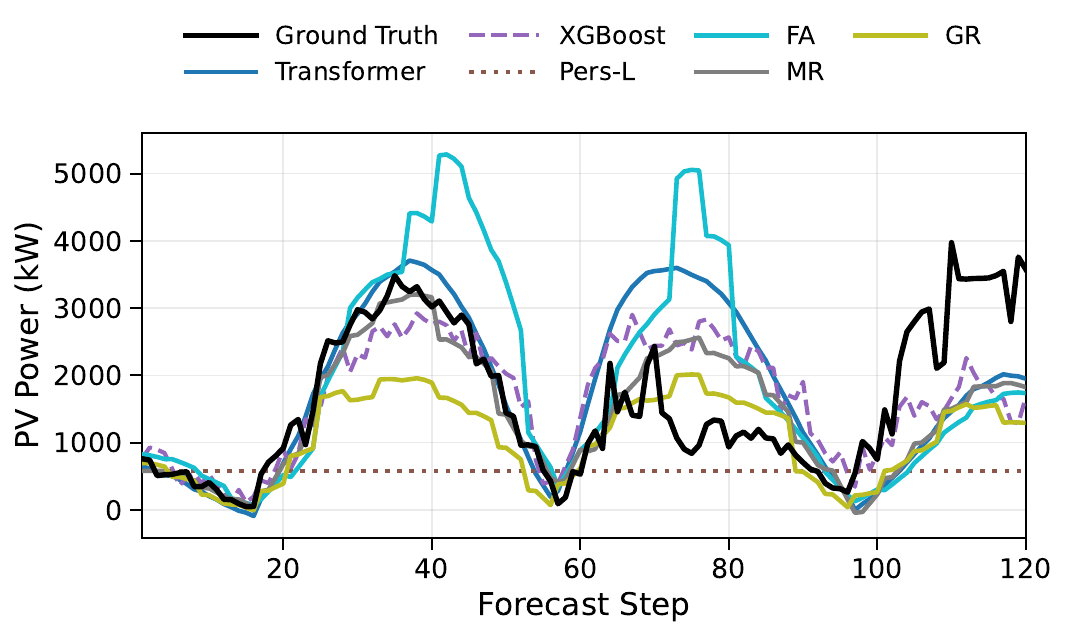}
        \label{fig:qual_case_h120}
    }
    \hfill
    \subfloat[$H = 7$ days]{
        \includegraphics[width=0.48\textwidth]{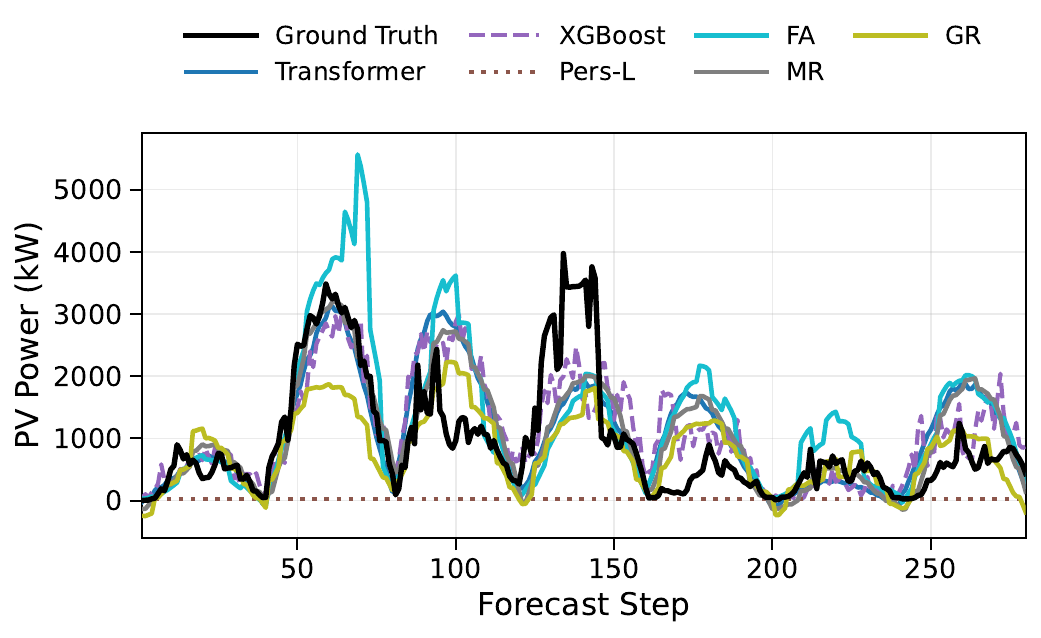}
        \label{fig:qual_case_h280}
    }

\caption{Representative zero-shot forecasting trajectories on unseen PV stations under different prediction horizons. The four subplots compare the proposed variants with conventional station-specific baselines. While most methods capture the general generation trend at short horizons, larger discrepancies appear at longer horizons, particularly around rapid ramps and multi-peak patterns.}
    
    \label{fig:qualitative_cases}
\end{figure*}

We first evaluate the forecasting performance on the 139 held-out PV stations. The shared models are trained using only the 340 development stations and are directly applied to the held-out stations without station-specific retraining. Results are reported for four forecasting horizons: 6 hours, 1 day, 3 days, and 7 days.

\subsubsection{Comparison with Shared and Generalist Models}

Table~\ref{tab:global_results} compares the proposed models with the shared Transformer and generalist time-series baselines. MR achieves the lowest MAE and RMSE at all four forecasting horizons. Compared with the shared Transformer, its gains are more evident at short and medium horizons, while the difference becomes small at 7 days.

The three pretraining objectives show a consistent ranking under the standard forecasting setting, with MR performing best, followed by FA and GR. This indicates that combining forecast-aligned reconstruction with additional random reconstruction provides a stronger pretraining signal than using either objective alone.

The generalist baselines represent two different settings. Chronos and Chronos-MV are classic foundation-model baselines retrained on the development dataset, whereas TimesFM-3 uses large-scale pretrained parameters learned from a broad collection of real-world and synthetic time series. TimesFM-3 gives the strongest performance among the generalist models, but MR achieves lower MAE and RMSE across all horizons. The performance gap also widens as the forecasting horizon increases, particularly for the 3-day and 7-day forecasts. These results show that large-scale general-purpose pretraining does not automatically translate into dominant performance for distributed PV forecasting, while domain-specific pretraining remains particularly effective at longer forecasting horizons.

\subsubsection{Comparison with Station-Specific Models}

Table~\ref{tab:stationwise_results} compares the zero-shot models with conventional station-specific methods. For the learnable station-specific baselines, separate models are trained for each target station and forecasting horizon using historical data from that station. The proposed models, in contrast, are applied directly to the held-out stations without parameter updating.

MR remains competitive with, and in most cases outperforms, the locally trained baselines. At the 6-hour horizon, MR achieves an RMSE of 0.129, compared with 0.136 for the station-specific Transformer and 0.137 for Informer. At the 1-day and 3-day horizons, MR obtains RMSE values of 0.135 and 0.147, respectively, both lower than the corresponding results of the compared station-specific models. At the 7-day horizon, MR reaches 0.168, compared with 0.169 for Informer and 0.172 for the station-specific Transformer.

The comparison is particularly relevant for newly introduced PV stations. The station-specific models benefit from target-station historical data and require a separate training procedure for each station and horizon. MR reaches comparable or better accuracy without this retraining step. A single pretrained model can therefore be reused across held-out stations and forecasting horizons while maintaining forecasting accuracy close to, or better than, locally customized models.

\subsection{Station-Level Generalization Analysis}
\label{sec:generalization_analysis}

Beyond the averaged results in Sec.~\ref{sec:zero_shot_results}, we further examine how forecasting performance varies across individual held-out stations. Fig.~\ref{fig:rmse_boxplots_all} shows the station-level RMSE distributions of MR and representative station-specific baselines under the four forecasting horizons.

MR maintains a low median RMSE across all horizons and shows a relatively compact error distribution. The advantage is more evident for the 6-hour, 1-day, and 3-day forecasts, while the distributions become closer at the 7-day horizon. This trend is consistent with the averaged results in Table~\ref{tab:stationwise_results}. More importantly, the station-level distributions show that the overall performance of MR is not determined by a small number of favorable stations, but is observed across a broad range of the 139 held-out PV systems.

Representative zero-shot forecasts are further shown in Fig.~\ref{fig:qualitative_cases}. The four panels present representative held-out cases under different forecasting horizons and compare the prediction behavior of the proposed variants with conventional station-specific baselines. At short horizons, most methods reproduce the overall PV generation profile with relatively small deviations. Larger differences appear at the 3-day and 7-day horizons, particularly around peak generation periods and rapid changes in the power trajectory.

Among the three pretraining objectives, GR tends to underestimate several generation peaks, whereas FA shows larger deviations in some long-horizon periods. MR generally follows the observed trajectory more closely and provides a more balanced prediction across the four horizons. These examples are consistent with the quantitative results and illustrate how the mixed reconstruction objective behaves when transferred directly to an unseen PV station.

\begin{table*}[!t]
\renewcommand{\arraystretch}{1.15}
\caption{Forecasting performance under degraded future NWP availability on 139 held-out PV stations. Results are averaged over all test stations. Lower values indicate better performance.}
\label{tab:weather_missing_multi}
\centering
\setlength{\tabcolsep}{4pt}
\begin{tabular}{llcccccccc}
\hline
& & \multicolumn{2}{c}{6 h}
& \multicolumn{2}{c}{1 day}
& \multicolumn{2}{c}{3 days}
& \multicolumn{2}{c}{7 days} \\
\cline{3-4}\cline{5-6}\cline{7-8}\cline{9-10}
\textbf{Scenario} & \textbf{Method}
& MAE & RMSE
& MAE & RMSE
& MAE & RMSE
& MAE & RMSE \\
\hline

\multirow{3}{*}{Full forecast}
& FA & 0.100 & 0.137 & 0.106 & 0.145 & 0.116 & 0.156 & 0.133 & 0.178 \\
& MR & \textbf{0.095} & \textbf{0.129} & \textbf{0.100} & \textbf{0.135} & \textbf{0.111} & \textbf{0.147} & \textbf{0.127} & \textbf{0.168} \\
& GR & 0.107 & 0.140 & 0.116 & 0.150 & 0.130 & 0.165 & 0.143 & 0.180 \\
\hline

\multirow{3}{*}{\shortstack{Partial NWP\\(30\% masked)}}
& FA & 0.111 & 0.148 & 0.118 & 0.157 & 0.127 & 0.167 & 0.142 & 0.187 \\
& MR & \textbf{0.097} & \textbf{0.131} & \textbf{0.102} & \textbf{0.137} & \textbf{0.113} & \textbf{0.149} & \textbf{0.129} & \textbf{0.170} \\
& GR & 0.113 & 0.146 & 0.122 & 0.157 & 0.137 & 0.172 & 0.148 & 0.185 \\
\hline

\multirow{3}{*}{No forecast}
& FA & 0.145 & 0.187 & 0.156 & 0.200 & 0.170 & 0.216 & 0.180 & 0.228 \\
& MR & \textbf{0.123} & \textbf{0.169} & \textbf{0.140} & \textbf{0.186} & \textbf{0.164} & \textbf{0.210} & \textbf{0.174} & 0.219 \\
& GR & 0.140 & 0.181 & 0.152 & 0.194 & 0.170 & 0.214 & \textbf{0.175} & \textbf{0.218} \\
\hline

\end{tabular}
\end{table*}

\begin{figure*}[!t]
    \centering
    \subfloat[Partial forecast availability]{
        \includegraphics[width=0.48\textwidth]{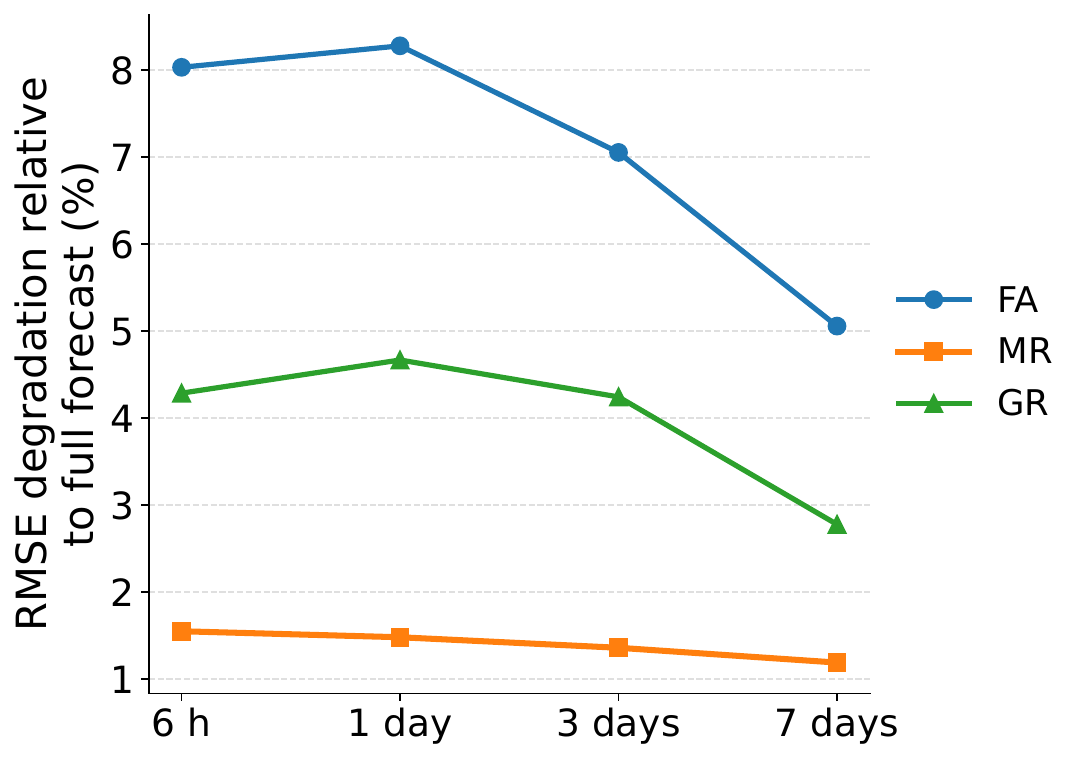}
        \label{fig:weather_missing_partial}
    }
    \hfill
    \subfloat[No future forecast]{
        \includegraphics[width=0.48\textwidth]{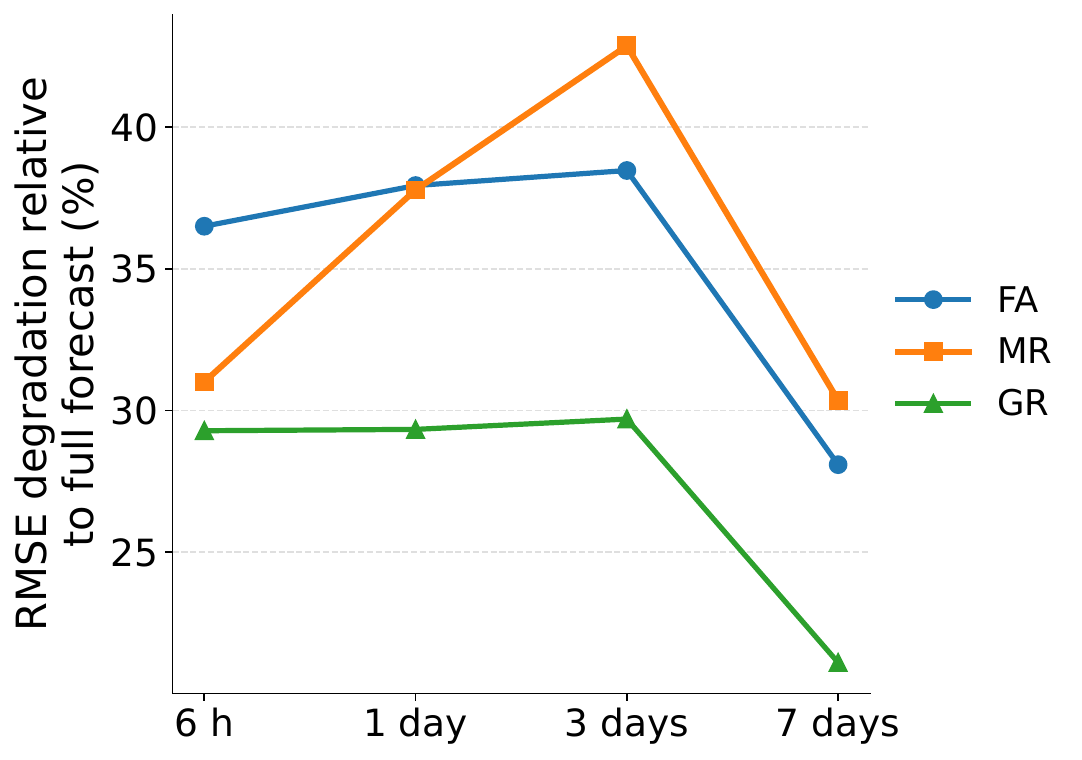}
        \label{fig:weather_missing_none}
    }
    \caption{Relative RMSE degradation under degraded future NWP availability. Degradation is computed relative to the full-forecast setting. Lower values indicate stronger robustness to reduced or unavailable future NWP inputs.}
    \label{fig:weather_missing_degradation}
\end{figure*}

\subsection{Robustness to Degraded NWP Availability}
\label{sec:nwp_robustness}

The previous experiments consider the standard forecasting setting in which the complete future NWP sequence is available at the forecast origin. In practical operation, however, NWP inputs may be incomplete because of delayed updates, communication failures, or missing forecast records. To examine the effect of reduced NWP availability, we evaluate FA, GR, and MR under three settings: full forecast, partial NWP availability with 30\% patch missingness, and no future forecast. Table~\ref{tab:weather_missing_multi} reports the absolute forecasting errors, while Fig.~\ref{fig:weather_missing_degradation} shows the corresponding relative RMSE degradation with respect to the full-forecast setting.

Reducing future NWP availability increases the forecasting error of all three models, confirming that future meteorological forecasts provide important information for PV prediction. The extent of this degradation, however, differs considerably among the three pretraining objectives.

Under partial NWP availability, MR is only slightly affected across all four forecasting horizons. Its RMSE increases from 0.129 to 0.131 at 6 hours, from 0.135 to 0.137 at 1 day, from 0.147 to 0.149 at 3 days, and from 0.168 to 0.170 at 7 days. Accordingly, Fig.~\ref{fig:weather_missing_degradation} shows a relative RMSE degradation of only about 1--2\% for MR, compared with larger changes for FA and GR. This behavior is consistent with the mixed reconstruction objective: in addition to forecasting-aligned supervision, MR is exposed to randomly masked variables during pretraining and therefore learns to reconstruct missing information from different parts of the available temporal and meteorological context.

When future NWP is completely unavailable, all three models experience a substantial loss of accuracy. GR shows the smallest \emph{relative} RMSE degradation across the four forecasting horizons. At the 3-day horizon, for example, its RMSE increases from 0.165 to 0.214, corresponding to a degradation of approximately 30\%, compared with about 38\% for FA and 43\% for MR. This stronger relative tolerance is consistent with the generic reconstruction objective of GR, which exposes the model to different partial-observation patterns during pretraining and reduces its dependence on a fixed set of available variables. However, GR starts from a higher error under the full-NWP setting, and its smaller relative degradation therefore does not translate into the best absolute forecasting accuracy. MR still achieves the lowest MAE and RMSE across most horizons after future NWP is removed.

These results reveal different characteristics of the three pretraining objectives. FA is closely aligned with the standard forecasting setting and therefore benefits strongly from the availability of future NWP. GR is less sensitive to the complete removal of NWP, but its generic reconstruction objective results in lower forecasting accuracy when full meteorological information is available. MR combines forecasting-oriented and random reconstruction supervision, providing the best absolute accuracy under normal and partially degraded NWP conditions while retaining reasonable performance when future NWP is completely unavailable.

\subsection{Downstream Adaptation to Region-Specific Operational Objectives}
\label{sec:downstream_adaptation}

The previous experiments evaluate direct zero-shot transfer without using target-station data for parameter updating. We next consider a downstream adaptation setting in which historical data from a target station are available and the pretrained model is further optimized for region-specific operational objectives.

For each target station, fine-tuning starts from the same pretrained checkpoint and is performed independently using the historical data of that station. All model parameters remain trainable during adaptation. The training, validation, and test periods follow the same chronological split defined in Sec.~\ref{sec:dataset_protocol}. Fine-tuning uses a learning rate of $1\times10^{-5}$ for up to 30 epochs, with early stopping based on validation performance. Since the regional KPIs contain non-smooth operations that are not directly suitable for gradient-based optimization, KPI-aligned differentiable surrogate losses are used during fine-tuning, including p.u. MSE for Shandong, soft-gated normalized MSE for Nanfang, and softmax-weighted MSE for Huabei. The final KPI values are computed using the original definitions in Sec.~\ref{sec:evaluation}.

Table~\ref{tab:kpi_finetune_summary} summarizes the adaptation results for the 1-day and 3-day forecasting horizons. A large majority of stations obtain positive KPI changes after fine-tuning, and the proportion of improved stations exceeds 90\% in almost all settings. This shows that the shared pretrained model can provide a useful initialization when target-station data become available and the forecasting objective changes from generic prediction accuracy to a specific operational assessment rule.

The final KPI results also reveal differences among the three pretraining objectives. MR achieves the highest final KPI in Huabei and Shandong for both forecasting horizons, reaching 79.60\% and 86.81\%, respectively, for the 1-day setting and 78.35\% and 85.54\% for the 3-day setting. GR gives slightly higher final scores in Nanfang, reaching 57.02\% at 1 day and 53.61\% at 3 days. These results show that no single pretraining objective dominates every regional criterion, although MR provides consistently strong initialization across the three operational settings.

GR exhibits larger average KPI improvements in several cases. This should be interpreted together with its lower zero-shot forecasting accuracy, which leaves more room for improvement during station-specific fine-tuning. The final KPI score therefore provides a more direct measure of the adapted model than the magnitude of the improvement alone.

Overall, the adaptation results complement the zero-shot experiments in Sec.~\ref{sec:zero_shot_results}. The pretrained model can either be deployed directly at an unseen station or further adapted when local historical data and region-specific operational objectives are available.

\begin{table*}[!t]
\renewcommand{\arraystretch}{1.12}
\caption{Downstream adaptation to region-specific operational objectives under the 1-day and 3-day forecasting settings. Results are averaged over 139 held-out PV stations. Bold values indicate the highest final Mean KPI for each forecasting horizon and regional criterion.}
\label{tab:kpi_finetune_summary}
\centering
\setlength{\tabcolsep}{5pt}
\scriptsize
\begin{tabular}{llcccc}
\hline
\textbf{Horizon} & \textbf{Method} & \textbf{Region} &
\textbf{Improved stations (\%)} &
\textbf{Mean $\Delta$KPI (pp)} &
\textbf{Mean KPI (\%)} \\
\hline

\multirow{9}{*}{1 day}
& FA & Huabei   & 97.84  & 0.98  & 77.66 \\
&    & Nanfang  & 99.28  & 7.44  & 52.52 \\
&    & Shandong & 90.65  & 0.20  & 85.74 \\
\cline{2-6}

& MR & Huabei   & 98.56  & 1.03  & \textbf{79.60} \\
&    & Nanfang  & 100.00 & 8.67  & 56.86 \\
&    & Shandong & 99.28  & 0.29  & \textbf{86.81} \\
\cline{2-6}

& GR & Huabei   & 99.28  & 1.71  & 79.54 \\
&    & Nanfang  & 100.00 & 17.31 & \textbf{57.02} \\
&    & Shandong & 98.56  & 1.70  & 86.75 \\
\hline

\multirow{9}{*}{3 days}
& FA & Huabei   & 99.28  & 1.29  & 76.54 \\
&    & Nanfang  & 100.00 & 8.74  & 49.72 \\
&    & Shandong & 93.53  & 0.25  & 84.65 \\
\cline{2-6}

& MR & Huabei   & 99.28  & 1.03  & \textbf{78.35} \\
&    & Nanfang  & 100.00 & 10.40 & 52.84 \\
&    & Shandong & 94.96  & 0.25  & \textbf{85.54} \\
\cline{2-6}

& GR & Huabei   & 98.56  & 1.74  & 78.06 \\
&    & Nanfang  & 100.00 & 18.01 & \textbf{53.61} \\
&    & Shandong & 99.28  & 1.92  & 85.45 \\
\hline
\end{tabular}

\vspace{2pt}
\footnotesize \textbf{Notes.}
``Improved stations (\%)'' denotes the percentage of stations with a positive KPI gain after adaptation.
``Mean $\Delta$KPI (pp)'' denotes the average KPI change from the pretrained zero-shot model to the adapted model, measured in percentage points.
``Mean KPI (\%)'' reports the final averaged KPI after adaptation under each regional assessment rule.
\end{table*}

% \section{Conclusion}

% This paper proposed a pretrained forecasting framework for scalable photovoltaic power forecasting in distributed multi-station scenarios. By reformulating PV forecasting as a sequence imputation problem with known-future meteorological inputs, the framework supports unified pretraining across multiple horizons and heterogeneous PV stations. Among the investigated strategies, the mixed reconstruction variant MR achieved the best overall performance.

% Experiments on a real-world dataset of 499 PV stations showed strong zero-shot cold-start transferability to unseen stations. MR consistently outperformed both unified global baselines and conventional station-specific models across multiple forecasting horizons, without requiring station-specific retraining. Additional results demonstrated favorable robustness under weather-missing scenarios and effective adaptation to region-specific KPI objectives through lightweight fine-tuning.

% Overall, the findings suggest that scalable PV forecasting benefits from cross-station shared learning together with task-aligned pretraining that explicitly leverages future weather dependency. Future work will investigate larger-scale pretraining across broader regions and continual adaptation under evolving operating conditions.

\section{Conclusion}

This paper presents a foundation-style pretraining framework for scalable distributed PV forecasting. A shared model is pretrained across multiple PV stations and forecasting horizons, while NWP-assisted forecasting is formulated as a sequence-completion task under the operational availability of historical observations and future weather forecasts. Among the three reconstruction objectives, MR achieves the best overall performance. Experiments on 479 distributed PV stations, including 340 development stations and 139 spatially held-out stations, show that MR provides strong zero-shot forecasting performance from 6 hours to 7 days and outperforms representative station-specific, shared, and generalist time-series baselines. Comparison with a recent large-scale foundation model further indicates that broad general-purpose pretraining does not necessarily outperform domain-specific pretraining for distributed PV forecasting, particularly at longer horizons. The proposed framework also remains robust under degraded NWP availability and can be further adapted to region-specific operational objectives when local data become available. Future work will investigate cross-regional transfer, probabilistic forecasting, and the integration of pretrained PV forecasts with downstream power-system decision-making.

\section*{Acknowledgment}

This work is supported by the RISUD Projects (No. P0042845) and the RISE Project (No. P0051003) of The Hong Kong Polytechnic University, International Centre of Urban Energy Nexus of The Hong Kong Polytechnic University (No. P0047700), and KKS Synergy project, Energy flexibility through synergies of big data, novel technologies \& systems, and innovative markets (202000739). This work is financially supported by The Major Science and Technology Projects of Ningbo (No. 2022Z236), Natural Science Foundation of Ningbo of China (No. 2023J027), as well as by the High Performance Computing Centers at Eastern Institute of Technology, Ningbo, and Ningbo Institute of Digital Twin.

% Use the style of numbering in square brackets.
% If nothing is used, default style will be taken.
%\begin{enumerate}[a)]
%\item 
%\item 
%\item 
%\end{enumerate}  

%\begin{itemize}
%\item 
%\item 
%\item 
%\end{itemize}  

%\begin{description}
%\item[]
%\item[] 
%\item[] 
%\end{description}  

% \clearpage %%Remove this from your manuscript

% % Figure
% \begin{figure}%[]
%   \centering
% %    \includegraphics{}
%     \caption{}\label{fig1}
% \end{figure}

% \begin{table}%[]
% \begin{tabular*}{\tblwidth}{@{}LL@{}}
% \toprule
% \midrule
%  & \\
%  & \\
%  & \\
%  & \\
% \bottomrule
% \end{tabular*}
% \end{table}

% Uncomment and use as the case may be
%\begin{theorem} 
%\end{theorem}

% Uncomment and use as the case may be
%\begin{lemma} 
%\end{lemma}


\printcredits

% \bibliographystyle{cas-model2-names}

\bibliography{cas-refs}



\end{document}